\documentclass[conference]{IEEEtran}
\IEEEoverridecommandlockouts
\usepackage{cite}
\usepackage{amsmath,amssymb,amsfonts}
\usepackage{algorithm}
\usepackage{algorithmic}
\usepackage{graphicx}
\usepackage{textcomp}
\usepackage{xcolor}
\def\BibTeX{{\rm B\kern-.05em{\sc i\kern-.025em b}\kern-.08em
    T\kern-.1667em\lower.7ex\hbox{E}\kern-.125emX}}
\begin{document}

\title{A Riemannian Primal-dual Algorithm Based on Proximal Operator and its Application in Metric Learning\\
}

\author{\IEEEauthorblockN{1\textsuperscript{st} Shijun Wang}
\IEEEauthorblockA{\textit{Dept. of Artificial Intelligence} \\
\textit{Ant Financial Services Group}\\
Seattle, USA \\
shijun.wang@alibaba-inc.com}
\and
\IEEEauthorblockN{2\textsuperscript{nd} Baocheng Zhu}
\IEEEauthorblockA{\textit{Dept. of Artificial Intelligence} \\
\textit{Ant Financial Services Group}\\
Shanghai, China \\
baocheng.zbc@antfin.com}
\and
\IEEEauthorblockN{3\textsuperscript{rd} Lintao Ma}
\IEEEauthorblockA{\textit{Dept. of Artificial Intelligence} \\
\textit{Ant Financial Services Group}\\
Shanghai, China \\
lintao.mlt@antfin.com}
\and
\IEEEauthorblockN{4\textsuperscript{th} Yuan Qi}
\IEEEauthorblockA{\textit{Dept. of Artificial Intelligence} \\
\textit{Ant Financial Services Group}\\
Hangzhou, China \\
yuan.qi@antfin.com}
}

\maketitle

\begin{abstract}
In this paper, we consider optimizing a smooth, convex, lower semicontinuous
 function in Riemannian space with constraints.
 To solve the problem, we first convert it to a dual problem and then propose
 a general primal-dual algorithm to optimize the primal and dual variables
 iteratively.
 In each optimization iteration, we employ a proximal operator to search
 optimal solution in the primal space.
 We prove convergence of the proposed algorithm and show its non-asymptotic
 convergence rate.
 By utilizing the proposed primal-dual optimization technique, we propose
 a novel metric learning algorithm which learns an optimal feature transformation matrix in the Riemannian space of positive definite matrices.
 Preliminary experimental results on an optimal fund selection problem in
 fund of funds (FOF) management for quantitative investment showed its efficacy.
\end{abstract}

\section{Introduction}
Many machine learning problems can be solved by optimization algorithms
 which minimize or maximize a predefined objective function under certain
 constraints if there are any.
 In the past decades, searching optimal variables in Euclidean space is
 a mainstream direction for optimization techniques.
 In recent years, there is a shift from Euclidean space to Riemannian space
 due to manifold structures existed in many machine learning problems\cite{nishimori2006riemannian,absil2007trust,Vandereycken2013,cheng2013riemannian,
 cherian2017riemannian,absil2010optimization}.
 To solve optimization problems in the Riemannian space, a straightforward
 method is to generalize optimization algorithms developed in the Euclidean
 space to the Riemannian space with consideration of manifold constraints on the variables to be optimized. Gradient descent methods, Newton's methods and conjugate gradient methods can be natural extended from the Euclidean space to the Riemannian space, see  \cite{gabay1982minimizing, smith1994optimization, BentoFO12, sato2015new} and references therein. Studies on Riemannian accelerated gradient methods, quasi-Newton algorithms like BFGS and adaptive optimization methods can be found in \cite{liu2017accelerated, qi2010riemannian,abs-1810-00760}.
 P. -A. Absil et al. proposed a trust-region approach for optimizing a smooth function on a
 Riemannian manifold in which the trust-region subproblems are solved using
 a truncated conjugate gradient algorithm \cite{absil2007trust}.
 Furthermore, Qi and Agarwal generalized the adaptive regularization with cubics algorithm to Riemannian manifold, and obtain an upper bound on the iteration complexity which is optimal compared to the complexity of steepest descent and trust-region methods \cite{qi2011numerical,agarwal2018adaptive}.
 In recent years, variance reduction techniques drew tremendous attention
 for optimizing finite-sum problems\cite{schmidt2017minimizing,johnson2013accelerating,defazio2014saga}.
 Extending the idea of variance reduction for optimizing finite sums of geodesically smooth functions on Riemannian manifolds can be found in \cite{zhang2016riemannian,sato2017riemannian,kasai2017riemannian}. 

Although intensive studies on Riemannian manifold optimization, all the above works have only considered
problems on unconstrained manifold, the only constraint is that the solution has to lie on the manifold. This severely limits the scope of possible applications with those methods. In many problems, additional equality or inequality constraints need to be imposed. There are few works addressed this problem, Hauswirth et al. extended the projected gradient descent algorithm to Riemannian manifold with inequality constraints and show its well-behaved convergent behaviour, but the manifold is restricted to submanifold of Euclidean space \cite{hauswirth2016projected}. 
Zhang et al. developed an ADMM-like primal-dual approach to solve nonconvex and nonsmooth multi-block optimization over Riemannian manifold with coupled linear equality constraints\cite{zhang2017primal}.

In this paper, we consider the following general nonlinear primal problem
 which is constrained on a complete Riemannian manifold $M$:
\begin{equation}
\underset{x\in M}{min}f\left(x\right),\label{eq:1}
\end{equation}
subject to constraints: $h\left(x\right)\preceq0$, where $f\in C^{2}\left(M,R\right)$ and $h=\left(h_{1},h_{2},...,h_{m}\right)\in C^{2}\left(M,R\right)$ are closed proper, convex, lower semicontinuous (l.s.c.) and real-valued functions. A function is of class $C^{2}$ if its first and second derivatives both exist and are continuous.

\subsection{Previous Works}
Khuzani and Li studied stochastic primal-dual method on the Riemannian manifolds with bounded sectional curvature\cite{khuzani2017stochastic}. They proved non-asymptotic convergence of the primal-dual method and established a connection between convergence rate and sectional curvature lower bound.
 In their algorithm, standard gradient descent method followed by exponential
 map to search optimal variable in each iteration.
 In recent years, proximal algorithms emerge from many machine learning
 applications due to their capability on handling nonsmooth, constrained,
 large-scale or distributed problems\cite{rockafellar1976augmented,parikh2014proximal}.
 Ferreira and Oliveira considered minimization problem on a Riemannian manifold with nonpositive sectional curvature. They solved the problem by extending the proximal method in the Euclidean space to the Riemannian space \cite{ferreira2002proximal}. Recent advances on Riemannian proximal point method can be found in \cite{bavcak2013proximal,chen2018proximal}.

\subsection{Our Contributions}
Previous works concentrate on either constrained Euclidean space optimization or unconstrained Riemannian space optimization, we propose a novel algorithm to solve constrained optimization problem (\ref{eq:1}) over the Riemannian manifold. We first convert it to a dual problem and then use a general primal-dual
 algorithm to optimize the primal and dual variables iteratively.
 In each optimization iteration, we employ the proximal point algorithm and gradient ascend method alternatively
 to search optimal solution in the primal and dual space. 
 We prove convergence of the proposed algorithm and show its non-asymptotic convergence rate.
 To show the efficacy of the proposed primal-dual algorithm for optimizing nonlinear l.s.c. functions in $C^{2}$ space with constraints, we propose a novel metric learning algorithm and
 solved it using the proposed Riemannian primal-dual method.

\section{Notation and Preliminaries}
Let $M$ be a connected and finite dimensional manifold with dimensionality of $m$.
 We denote by $T_{p}M$ the tangent space of $M$ at $p$.
 Let $M$ be endowed with a Riemannian metric $\left\langle .,.\right\rangle $, with corresponding norm denoted by $\parallel.\parallel$, so that $M$ is now a Riemannian manifold\cite{eisenhart2016riemannian}.
 We use $l\left(\gamma\right)=\int_{a}^{b}\parallel\gamma^{\prime}\left(t\right)\parallel dt$ to denote the length of a piecewise smooth curve $\gamma:\left[a,b\right]\longrightarrow M$ joining $x^{\prime}$ to $x$, i.e., such that $\gamma\left(a\right)=x^{\prime}$ and $\gamma\left(b\right)=x$.
 Minimizing this length functional over the set of all piecewise smooth curves passing $x^{\prime}$ and $x$ we get a Riemannian distance $d\left(x^{\prime},x\right)$ which induces the original topology on $M$.
 Take $x\in M,$ the exponential map $exp_{x}:T_{x}M\longrightarrow M$ is defined by $exp_{x}v=\gamma_{v}\left(1,x\right)$ which maps a tangent vector $v$ at $x$ to $M$ along the curve $\gamma$.
 For any $x^{\prime}\in M$ we define the exponential inverse map $exp_{x^{\prime}}^{-1}:M\longrightarrow T_{x^{\prime}}M$
which is $C^{\infty}$ and maps a point $x^{\prime}$ on $M$ to a tangent vector at $x$ with $d\left(x^{\prime},x\right)=\parallel exp_{x^{\prime}}^{-1}x\parallel$.
 We assume $\left(M,d\right)$ is a complete metric space, bounded and all closed subsets of $M$ are compact.
 For a given convex function $f:M\rightarrow R$ at $x^{\prime}\in M$, a vector $s\in T_{x^{\prime}}M$ is called subgradient of $f$ at $x^{\prime}\in M$ if $f\left(x\right)\geq f\left(x^{\prime}\right)+<s,exp_{x^{\prime}}^{-1}x>$, for all $x\in M$.
 The set of all subgradients of $f$ at $x^{\prime}\in M$ is called subdifferential of $f$ at $x^{\prime}\in M$ which is denoted by $\partial f\left(x^{\prime}\right)$.
 If $M$ is a Hadamard manifold which is complete, simply connected and has everywhere non-positive sectional curvature, the subdifferential of $f$ at any point on $M$ is nonempty\cite{ferreira2002proximal}.

\section{The Algorithm}
By employing duality, we convert original optimization problem (\ref{eq:1}) to an augmented Lagrangian function (generic saddle-point problem):

\begin{equation}
L\left(x,\lambda\right)=f\left(x\right)+<\lambda,h\left(x\right)>-\frac{\alpha}{2}\parallel\lambda\parallel^{2},\label{eq:2}
\end{equation}
where 
$\lambda\in\mathbb{\mathbb{R}}_{+}^{m}$ is Lagrangian dual vector for inequality constraints, and $\alpha>0$ is a regularization parameter which weights norm of the dual variables. The norm of the Lagrangian dual variables $\parallel\lambda\parallel$ is upper bounded and so is gradient of the deterministic Lagrangian function $L\left(x_{t},\lambda_{t}\right)$.

To solve the generic primal-dual problem shown in eq. (\ref{eq:2}), we propose a primal-dual algorithm based on proximal operator (see Algorithm 1).

\begin{algorithm}
\caption{Riemannian Primal-dual Algorithm Based on Proximal Operator}
\begin{algorithmic} 
\STATE 1.initialize: set initial point 
$x_{0}\in M$, 
$\lambda_{0}=0$
, and step size sequence 
$\left\{ \eta_{t}\right\} _{t=0}^{T}$
 which is decreasing and 
$\eta_{t}\in\mathbb{R}_{+}$
.
\STATE 2. for 
$t=0,1,2,...,T$
 do:
\begin{align} 
x_{t+1}&=prox_{L}\left(x_{t}\right)=arg\underset{x\in M}{min}\left\{ L\left(x,\lambda_{t}\right)+\frac{1}{2\eta_{t}}d^{2}\left(x_{t},x\right)\right\} , \label{eq:3} \\
\lambda_{t+1}&=\left[\lambda_{t}+\eta_{t}grad_{\lambda_{t}}L\left(x_{t+1},\lambda_{t}\right)\right]_{+},\label{eq:4}
\end{align}
where 
$grad_{\lambda}L\left(x_{t+1},\lambda\right)=h\left(x_{t+1}\right)-\alpha\lambda$, and $\left[x\right]_{+}$ means for a vector $x$, change every negative entry of $x$ to zero.

\end{algorithmic}
\end{algorithm}

\textbf{Assumption 1}.
 Assume $M$ is a compact Riemannian manifold with finite diameter 
$R=sup_{x,y\in M}d\left(x,y\right)$
 and non-positive sectional curvature.
 For any 
$x\in M$, the following gradients are bounded by 
$\parallel grad_{x}f\left(x\right)\parallel\leq C$, 
$\parallel grad_{x}h_{k}\left(x\right)\parallel\leq C$, and $\mid h_{k}\left(x\right)\mid\leq G$, $k=1,2,...,m$.

With Assumption 1, we have the following theorem.

\textbf{Theorem 1}. Assume problem (\ref{eq:3}) has a saddle-point 
$\left(x_{*},\lambda_{*}\right)$
 and Assumption 1 hold.
 Let 
$\left\{ x_{t}\right\} _{t=0}^{T-1}$
 be a finite sequence generated by Algorithm 1 iteratively, and step size $\eta_{t}\alpha\leq1$, 
$t\in[T]=\left\{ 0,1,2,...,T-1\right\} $.
 Then,
\begin{multline}
\underset{t\in\left[T\right]}{min}f\left(x_{t+1}\right)-f\left(x_{*}\right)\\
\leq\frac{1}{\sum_{t=0}^{T-1}\eta_{t}}\left(\frac{1}{2}d^{2}\left(x_{*},x_{0}\right)+2mG^{2}\sum_{t=0}^{T-1}\left(\eta_{t}^{2}\right)\right) \label{eq:5}
\end{multline}
for all 
$x_{t}\in M,t\in\left[T\right]$.

Proof of Theorem 1 is shown in appendix A.

From Theorem 1, we could derive the following corollary:

\textbf{Corollary 1}. (Non-asymptotic Convergence) By choosing step size 
$\eta_{t}=\frac{1}{\sqrt{t+1}}$, and 
$\alpha\eta_{t}\leq1$ for all 
$t\in[T]$, the sequence 
$\left\{ x_{t}\right\} $, 
$t\in[T]$ generated by Algorithm 1 converges at rate:
\begin{equation}
\underset{t\in\left[T\right]}{min}f\left(x_{t+1}\right)-f\left(x_{*}\right)=\mathcal{O}\left(\frac{log\left(T\right)}{\sqrt{T}-1}\right).\label{eq:6}
\end{equation}

Proof of Corollary 1 is shown in appendix B.

\section{Application in Metric Learning}
Metric learning is a technique to learn a distance metric in data feature space, and finds application in various machine learning tasks relying on distances or similarities measure, like classification, clustering, dimensionality reduction and domain adaptation, to name a few \cite{kulis2013metric,bellet2013survey,
xing2003distance,weinberger2006distance,davis2007information,wang2009information,zadeh2016geometric,mahadevan2018unified}. 
Most methods learn the metric (positive definite matrix $\mathbf{W}$) in a weakly-supervised way from pairwise or triplet constraints of data points. In general, metric learning can be formulated as an optimization problem that shares the same form as standard regularized empirical risk minimization:  
\begin{equation}
\underset{\mathbf{W}}{min} \mathcal{L(\mathbf{W},\mathbf{X})} + \lambda\Omega(\mathbf{W}),\label{eq:7a}
\end{equation}
where $\mathbf{X}$ denotes training samples, $\mathcal{L}$ is the loss function associated with sample constraints and $\Omega$ is the regularizer, $\lambda$ is the trade-off parameter. Many methods are specified as a constrained optimization problem by writing down $\mathcal{L}$ explicitly as inequality constraints 
$h(\mathbf{W},\mathbf{X})\preceq 0$, although we can always transform it into an unconstrained problem using hinge loss or other tricks \cite{kulis2013metric,zadeh2016geometric}.

Some techniques are developed to solve metric learning optimization problem eq. (\ref{eq:7a}). Projected gradient descent and its stochastic version use traditional (stochastic) gradient descent followed by an orthogonal projection onto the positive semi-definite cone \cite{xing2003distance,levitin1966constrained,shalev2004online,jain2009online}.  
Bregman projections update based on one single constraint at each iteration, and perform a general non-orthogonal projection so that the chosen constraint is satisfied. After projecting, an appropriate correction is employed \cite{davis2007information}.

However, these methods do not fully use the intrinsic manifold structure of the problem, i.e. the learned metric must lie in a Riemannian space of positive definite matrices. So it is naturally an optimization problem on Riemannian manifold rather than Euclidean space. In this section, we apply the proposed method to metric learning problem and illustrate how to optimize a convex target function in a Riemannian manifold.

\subsection{Metric Learning Problem Formulation}
We consider the following convex metric learning problem with $\mathbf{W}$ in the Riemannian space 
$\mathbf{S}_{+}^{n}$
of 
$n\times n$
positive definite matrices:
\begin{equation}
\underset{\mathbf{W}\in\mathbf{S}_{+}^{n}}{min}\Omega\left(\mathbf{W}\right),\label{eq:7}
\end{equation}
s.t.

$\left(\mathbf{x}_{i}-\mathbf{x}_{j}\right)\mathbf{W}\left(\mathbf{x}_{i}-\mathbf{x}_{j}\right)^{\top}\leq u,\forall\left(i,j\right)\in C^{+},$

$\left(\mathbf{x}_{i}-\mathbf{x}_{j}\right)\mathbf{W}\left(\mathbf{x}_{i}-\mathbf{x}_{j}\right)^{\top}\geq l,\forall\left(i,j\right)\in C^{-},$
where 
$\Omega\left(\mathbf{W}\right)=\frac{1}{2}d^2(\mathbf{W},\mathbf{W}_{0})$, $d^2(\mathbf{W},\mathbf{W}_{0})=tr\left(\mathbf{W}\mathbf{W}_{0}^{-1}\right)-logdet\left(\mathbf{W}\mathbf{W}_{0}^{-1}\right)-n$ is the LogDet divergence which is a scale-invariant distance measure on Riemannian metrics manifold \cite{davis2007information}. $\mathbf{W}_{0}$ is a target transformation matrix initialized to identity (corresponds to the Euclidean distance) or inverse of data covariance matrix (corresponds to the Mahalanobis distance), $C^{+}/C^{-}$ is set of all sample pairs with the same/different labels. $u$ and $l$ are the upper/lower distance bound of similar/dissimilar pairs of points and are set to 5-th/95-th percentiles of the observed distribution of distances in the following experiments. It is known that space of all $n\times n$ positive definite Hermitian matrices is a Cartan-Hadamard manifold which is a simply connected complete Riemannian manifold with non-positive sectional curvature.
 
\subsection{Optimization by the Proposed Riemannian Primal-dual Algorithm}
By introducing relaxation variables $\mathbf{\xi}$, we have
\begin{equation}
\underset{\mathbf{W}\in\mathbf{S}_{+}^{n}}{min}\Omega\left(\mathbf{W}\right)+\frac{C_{1}}{2}\parallel\xi\parallel_{2}^{2}\text{,}\label{eq:8}
\end{equation}

s.t.

$\left(\mathbf{x}_{i}-\mathbf{x}_{j}\right)\mathbf{W}\left(\mathbf{x}_{i}-\mathbf{x}_{j}\right)^{\top}\leq u(1+\xi_{ij})$
, 
$\forall\left(i,j\right)\in C^{+}$
,

$\left(\mathbf{x}_{i}-\mathbf{x}_{j}\right)\mathbf{W}\left(\mathbf{x}_{i}-\mathbf{x}_{j}\right)^{\top}\geq l(1-\xi_{ij})$
, 
$\forall\left(i,j\right)\in C^{-}$.

Let's define 
$h_{+}\left(\mathbf{W}\right)=diag\left(\mathbf{X}_{+}\mathbf{W\mathbf{X}_{+}^{\top}}\right)-u(\mathbf{e}+\xi_{+})\leq0$ ($\mathbf{e}$ is a vector whose entries are all one, $diag\left(X\right)$ extracts diagonal elements of an input matrix $X$ and write them to a vector), and 
$h_{-}\left(\mathbf{W}\right)=-diag\left(\mathbf{X}_{-}\mathbf{W}\mathbf{X}_{-}^{\top}\right)+l(\mathbf{e}-\xi_{-})\leq0$
, where $\mathbf{X}_{+}/\mathbf{X}_{-}$ are matrices composed by sample pairs with the same / different labels (shape: number of samples by feature dimensions), and $\xi_{+}/\xi_{-}$ are corresponding relaxation vectors which are greater than or equal to zero. So we have
\begin{equation}
\underset{\mathbf{W}\in\mathbf{S}_{+}^{n}}{min}\Omega\left(\mathbf{W}\right)+\frac{C_{1}}{2}\parallel\xi_{+}\parallel_{2}^{2}+\frac{C_{1}}{2}\parallel\xi_{-}\parallel_{2}^{2},\label{eq:9}
\end{equation}
s.t.

$h_{+}\left(\mathbf{W}\right)=diag\left(\mathbf{X}_{+}\mathbf{W\mathbf{X}_{+}^{\top}}\right)-u(\mathbf{e}+\xi_{+})\leq0$,

$h_{-}\left(\mathbf{W}\right)=-diag\left(\mathbf{X}_{-}\mathbf{W}\mathbf{X}_{-}^{\top}\right)+l(\mathbf{e}-\xi_{-})\leq0$,

$-\xi_{+}\leq0$,

$-\xi_{-}\leq0$.

Further more, we define 
$h\left(\mathbf{W}\right)=\left[h_{+}\left(\mathbf{W}\right),h_{-}\left(\mathbf{W}\right)\right]^{\top}$
, and 
$\xi=\left[\xi_{+},\xi_{-}\right]$
, then
\begin{equation}
\underset{\mathbf{W}\in\mathbf{S}_{+}^{n}}{min}\Omega\left(\mathbf{W}\right)+\frac{C_{1}}{2}\parallel\xi\parallel_{2}^{2},\label{eq:10}
\end{equation}
s.t.

$h\left(\mathbf{W}\right)\leq0$,

$-\xi\leq0$.

By employing duality, we have the following augmented Lagrangian function:

$L\left(\mathbf{W},\xi,\lambda,\gamma\right)=\Omega\left(\mathbf{W}\right)+\frac{C_{1}}{2}\parallel\xi\parallel^{2}+<\lambda,h\left(\mathbf{W}\right)>+<\gamma,-\xi>-\frac{C_{2}}{2}\parallel\lambda\parallel^{2}-\frac{C_{2}}{2}\parallel\gamma\parallel^{2}$.

Now let's solve the above Lagrangian function using Algorithm 1. At each step $t+1$, we have the following updates:

$\mathbf{W}_{t+1}=arg\underset{\mathbf{W}\in\mathbf{S}_{+}^{n}}{min}\left\{ L\left(\mathbf{W},\xi,\lambda_{t},\gamma_{t}\right)+\frac{1}{2\eta_{t}}d^{2}\left(\mathbf{W}_{t},\mathbf{W}\right)\right\} ,$

$\mathbf{\xi}_{t+1}=arg\underset{\xi\geq0}{min}\left\{ L\left(\mathbf{W},\xi,\lambda_{t},\gamma_{t}\right)+\frac{1}{2\eta_{t}}d^{2}\left(\mathbf{\xi}_{t},\mathbf{\xi}\right)\right\} $,

$\lambda_{t+1}=\left[\lambda_{t}+\eta_{t}grad_{\lambda_{t}}L\left(\mathbf{W}_{t+1},\xi_{t+1},\lambda_{t},\gamma_{t}\right)\right]_{+}$, 

$\gamma_{t+1}=\left[\gamma_{t}+\eta_{t}grad_{\gamma_{t}}L\left(\mathbf{W}_{t+1},\xi_{t+1},\lambda_{t},\gamma_{t}\right)\right]_{+}$.

So,

$\mathbf{W}_{t+1}=$
$arg\underset{\mathbf{W}\in\mathbf{S}_{+}^{n}}{min}\{ \frac{1}{2}d^2(\mathbf{W},\mathbf{W}_{0})+<\lambda_{t},h\left(\mathbf{W}\right)>$
$+\frac{1}{2\eta_{t}}d^{2}\left(\mathbf{W}_{t},\mathbf{W}\right)\} $,

$\mathbf{\xi}_{t+1}=$
$arg\underset{\xi\geq0}{min}\{ \frac{C_{1}}{2}\parallel\xi\parallel^{2}+<\lambda_{t},h(\mathbf{W},\xi)>$
$+<\gamma_{t},-\xi>+\frac{1}{2\eta_{t}}d^{2}\left(\mathbf{\xi}_{t},\mathbf{\xi}\right)\}$,

$\lambda_{t+1}=\left[\lambda_{t}+\eta_{t}grad_{\lambda_{t}}\left[<\lambda_{t},h\left(\mathbf{W}_{t+1}\right)>-\frac{C_{2}}{2}\parallel\lambda_{t}\parallel^{2}\right]\right]_{+}$, 

$\gamma_{t+1}=\left[\gamma_{t}+\eta_{t}grad_{\gamma_{t}}\left[<\gamma_{t},-\xi_{t+1}>-\frac{C_{2}}{2}\parallel\gamma_{t}\parallel^{2}\right]\right]_{+}$.

In the following paragraphs, we will show how to update primal and dual variables in each iteration.

\textbf{(1)} We employ Riemannian gradient decent method to search optimal $\mathbf{W}_{t+1}$. Define

$J_{\mathbf{W}}=L\left(\mathbf{W},\xi,\lambda_{t},\gamma_{t}\right)=\frac{1}{2}d^2(\mathbf{W},\mathbf{W}_{0})+<\lambda_{t},h\left(\mathbf{W}\right)>$.

We have
$\mathbf{W}_{t+1}=R_{\mathbf{W}_{t}}\left(-\eta_{t}Grad_{\mathbf{W}_{t}}J_{\mathbf{W}_{t}}\right)$, where $Grad_{\mathbf{W}}J_{\mathbf{W}}=\frac{1}{2}\left(\mathbf{W}_{0}^{-1}-\mathbf{W}^{-1}\right)+<\lambda_{t},\frac{\partial h\left(\mathbf{W}\right)}{\partial \mathbf{W}}>$ is the Riemannian gradient and operator $R_W$ means retraction. See Appendix C for the full procedure.

\textbf{(2)}
 Define 
$J_{\xi}=\frac{C_{1}}{2}\parallel\xi\parallel^{2}+<\lambda_{t},h(\mathbf{W},\xi)>+<\gamma_{t},-\xi>+\frac{1}{2\eta_{t}}d^{2}\left(\mathbf{\xi}_{t},\mathbf{\xi}\right)$
, and 
$d^{2}\left(\mathbf{\xi}_{t},\mathbf{\xi}\right)=\parallel\xi-\mathbf{\xi}_{t}\parallel^{2}$
.

$grad_{\xi}J_{\xi}=C_{1}\xi-<\lambda_t,\binom{u}{l}>-\gamma_{t}+\eta_{t}\left(\xi-\mathbf{\xi}_{t}\right)$

$=\left(C_{1}+\eta_{t}\right)\xi-\eta_{t}\mathbf{\xi}_{t}-\gamma_{t}-<\lambda_t,\binom{u}{l}>=0,$

$\mathbf{\xi}_{t+1}=\left[\frac{1}{C_{1}+\eta_{t}}\left(\eta_{t}\mathbf{\xi}_{t}+\gamma_{t}+<\lambda_t,\binom{u}{l}>\right)\right]_{+}$.

\textbf{(3)}
$\lambda_{t+1}=\left[\left(1-C_{2}\eta_{t}\right)\lambda_{t}+\eta_{t}h\left(\mathbf{W}_{t+1}\right)\right]_{+}.$

\textbf{(4)}
$\gamma_{t+1}=\left[\left(1-C_{2}\eta_{t}\right)\gamma_{t}-\eta_{t}\xi_{t+1}\right]_{+}$
.

\subsection{Experimental Results}

Both investors and machine learning researchers showed great interests on
 applying machine learning to finance area in recent years.
 The Holy Grail of quantitative investment is selection of high-quality
 financial assets with good timing to achieve higher returns with less risk\cite{defusco2015quantitative}.
 To measure quality and trend of financial assets, technical, fundamental,
 and macroeconomic factors or features are developed, Usually multi-factor regression models\cite{fama2004capital} are deployed to find the most effective features to achieve higher asset return. However, when the number of features is large, or heterogeneity/multicollinearity exists in these features, traditional factor-oriented asset selection models tend to fail and may not give encouraging results. 

Each asset can be represented by a data point in high-dimensional feature space, then a good distance metric in the space is crucial for more accurate similarity measure between assets. In our treating, assets selection can be regarded as a classification problem, assets are divided into two groups, one with positive return and the other with negative return, the aim is to find an optimal distance metric in feature space to separate these two groups, which is exactly what eq. (\ref{eq:7}) formulated. Using metric learning approach to asset selection, above mentioned factor model problems can be largely alleviated. In following sections, we apply the proposed Riemannian primal-dual metric
 learning (RPDML) algorithm to fund of funds (FOF) management problem.
 FOF is a multi-manager investment strategy whose portfolios are composed
 by mutual or hedge funds which invest directly in stocks, bonds or other
 financial assets.

\subsection{Data}

In this research, we consider Chinese mutual funds that were publicly traded for at least 12 consecutive months in the period 2012-01-01 to 2018-12-31. We select a total of 697 funds with capital size larger than 100 million RMB.
 Fund features consist of totally 70 technical factors with different rolling
 windows (10, 14, 21, 28, 42, 63 and 90 trading days respectively), including
 ROC, EMA, MDD, STDDEV, Sharpe ratio, Sortino ratio, Calmar ratio, RSI,
 MACD, and Stability [https://www.investopedia.com/]. All features are normalized to have zero mean and unit variance.

\subsection{Backtest Protocol}

For mutual fund management, typical length of rebalancing interval is one
 quarter of a year.
 So we split original sequential fund data into segments of quarters.
 We use a rolling window prediction schema, in the training set, we learn a distance metric from 70 technical factors from previous quarter with quarterly return of current quarter as target. In the test set, features from current quarter are used to predict quarterly return of next quarter. To validate learned metric, a simple k-nearest neighbors (k-NN) algorithm is employed, the predict return of next quarter for each fund is based on the learned metric from training set. The hyper-parameter k of k-NN is set as 10. To avoid overfitting, rolling data before 2017-01-01 are used as validation and data from 2017-01-01 to 2018-12-31 as test. For convenience, following results are based on both validation and test set.

For each fund, with k nearest neighbor funds predicted, we use average of
 returns of the k neighbor funds in current quarter as prediction of the
 fund's return in the next quarter.
 At each quarterly rebalance day (the last trading day of each quarter), a top 10 buy trading strategy is used based on the prediction of each fund.
 In this strategy, we rank all funds based on their predicted quarterly
 return and select the top 10 funds for portfolio construction and rebalancing
 with equal weight.

We compare RPDML with the following four distance metrics and a baseline
 fund index: 

- Euclidean distance metric.

- Mahalanobis distance metric, which is computed as the inverse of covariance
 matrix of training samples.

- LMNN, a distance metric learning algorithm based on the large margin nearest
 neighbor classification \cite{weinberger2006distance}.

- ITML, a distance metric learning algorithm based on information geometry\cite{davis2007information}.

- GMML, a distance metric learning algorithm, the learned metric can be viewed as ``matrix geometric mean'' on
the Riemannian manifold of positive definite matrices \cite{zadeh2016geometric}.

- Baseline fund index, CSI Partial Equity Funds Index [http://www.csindex.com.cn/en
/indices/index-detail/930950]

\subsection{Result}
\begin{table*}[ht]
\centering
\caption{Spearman's correlation/Information coefficient for different algorithms.}
  \begin{tabular}{ |l|l|l|l|l|l|l| }
    \hline
    \textbf{Algorithm} & Euclidean & Mahalanobis & LMNN & ITML & GMML & RPDML \\ \hline
    \textbf{IC} & $0.018\pm 0.138$ & $0.050\pm 0.071$ & $0.017\pm 0.201$ & $0.030 \pm 0.071$ & $0.058 \pm 0.086$ & $\mathbf{0.069 \pm 0.187}$ \\ \hline
  \end{tabular}
  
\end{table*}

In the FoF setting, we care more about the order of predicted return than the absolute value. So we calculate the Spearman's rank-order correlation of the predicted return to the true return for different algorithms, a higher correlation means better predictive power. In financial community, this correlation is often called Information coefficient (IC). The calculation is done rollingly, and we show the mean and standard deviation of IC in Table 1. We can see that RPDML achieves highest mean correlation. The backtest performance using different prediction models is shown in Figure 1.
 We also show CSI Partial Equity Funds Index (dashed curves) as baseline
 fund index which reflects overall performance of all partial equity funds
 in China's financial market.
 From the top panel, we observed that all the experimental algorithms outperform
 baseline index, and among them, RPDML achieved the best performance, with
 total accumulated return of 148\% in the whole backtest period, while the
 worst portfolio with Euclidean distance metric only achieved 25\%, even less than CSI Partial Equity Funds Index. We also plot one year rolling maximum drawdown (MDD) in the bottom panel,
 MDD of the RPDML algorithm is quite low considering its superior accumulated return.

\begin{figure}[htbp]
\centerline{\includegraphics[width=1\columnwidth]{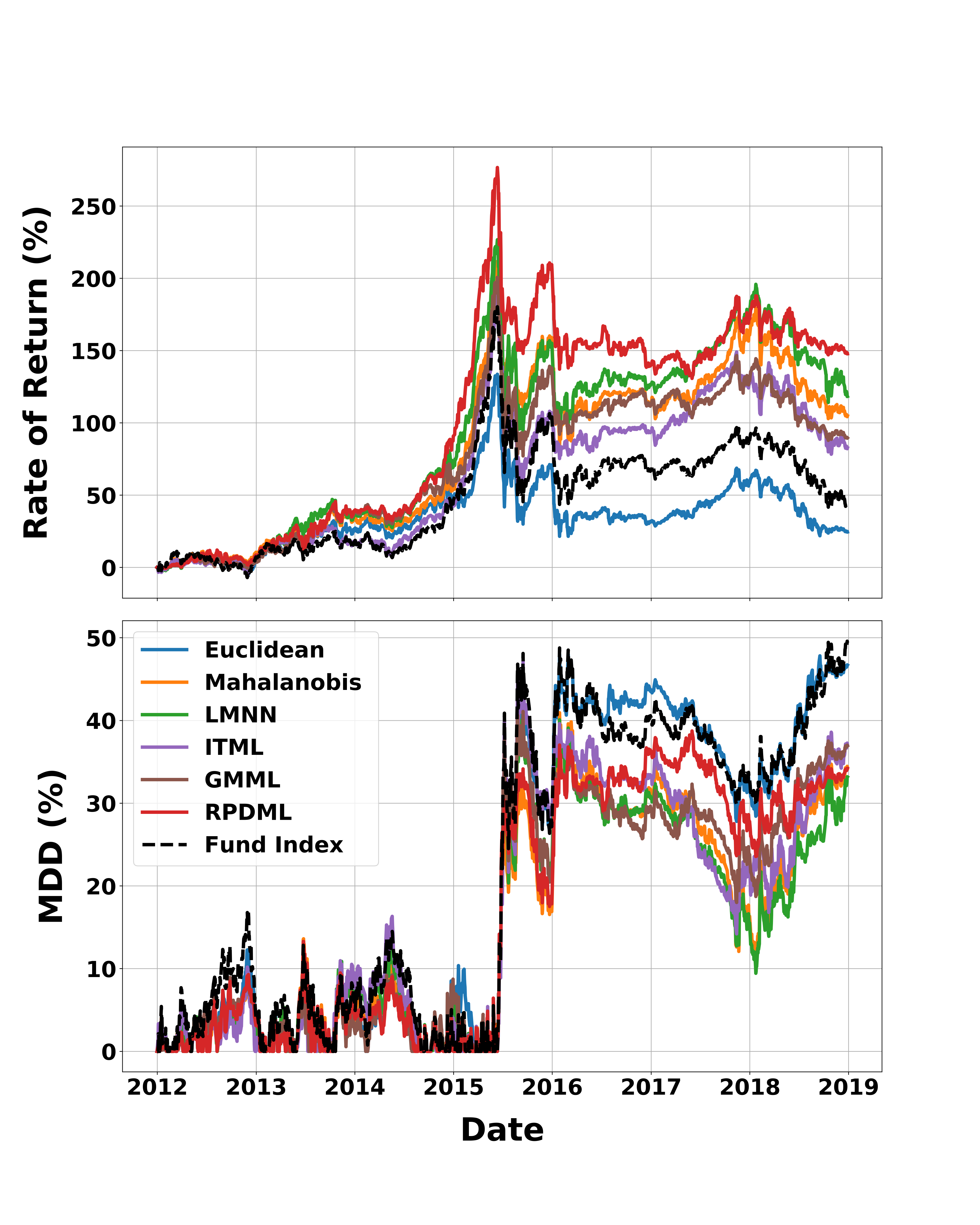}}
\caption{Portfolio performance comparison of each metric learning algorithm and a
 baseline fund index.
 (Top) Accumulate rate of return, (Bottom) Maximum drawdown.}
\label{fig}
\end{figure}

In Figure 2, we show annual returns of FOF of each algorithm.
 We can see that the proposed algorithm PRDML achieved highest returns in
 most years.
 Besides, we also notice that LMNN performed quite well in some
 years.

\begin{figure}[htbp]
\centerline{\includegraphics[width=1\columnwidth]{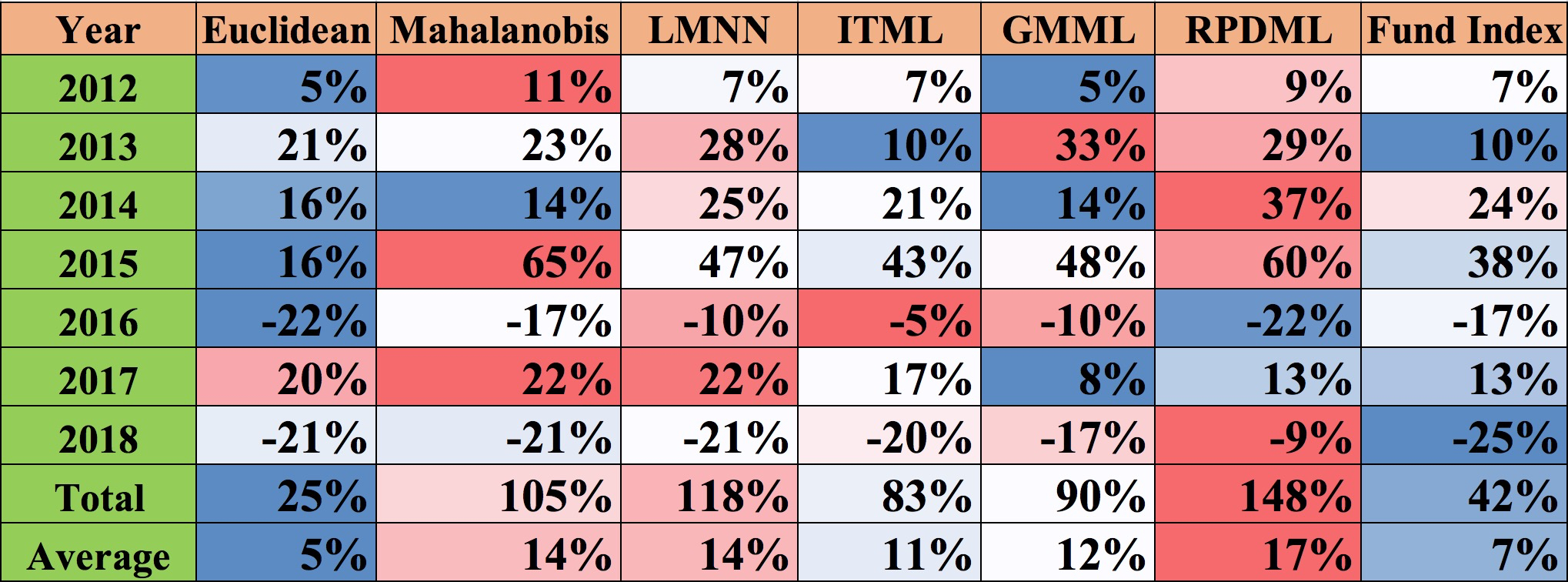}}
\caption{Annual return of FOF for each algorithm.}
\label{fig}
\end{figure}

\section{Conclusion}
In this paper, we propose a Riemannian primal-dual algorithm based on proximal
 operator for optimizing a smooth, convex, lower semicontinuous function
 on Riemannian manifolds with constraints.
 We prove convergence of the proposed algorithm and show its non-asymptotic
 rate.
 By utilizing the proposed primal-dual optimization technique, we propose
 a novel metric learning algorithm which learns an optimal feature transformation matrix in the Riemannian space of positive definite matrices.
 Preliminary experimental results on an optimal fund selection problem in
 FOF management for quantitative investment showed its efficacy.

\section{Appendix}

\subsection{Proof of Theorem 1}

Due to convexity of 
$L\left(x,\lambda\right)$, for any 
$x\in M$ we have
\begin{equation}
L\left(x,\lambda_{t}\right)\geq L\left(x_{t+1},\lambda_{t}\right)+<s,exp_{x_{t+1}}^{-1}x>,\label{eq:11}
\end{equation}

where 
$s\in\partial L\left(x_{t+1},\lambda_{t}\right)$ and 
$exp_{x_{t+1}}^{-1}x\in T_{x_{t+1}}M$.

Because $M$ is a Hadamard manifold which has non-positive curvature, then
\begin{equation}
d^{2}\left(x,x_{t}\right)\geq d^{2}\left(x,x_{t+1}\right)+d^{2}\left(x_{t+1},x_{t}\right)-2<exp_{x_{t+1}}^{-1}x_{t},exp_{x_{t+1}}^{-1}x>.\label{eq:12}
\end{equation}
 [ref.\cite{bento2017iteration}, Proposition 1]

Multiplying eq. (\ref{eq:12}) by 
$\frac{1}{2\eta_{t}}$ and summing the result with eq.
 (\ref{eq:11}), we get the following inequality:
\begin{equation}
\begin{split}
&L\left(x,\lambda_{t}\right)+\frac{1}{2\eta_{t}}d^{2}\left(x,x_{t}\right)\\
&\geq L\left(x_{t+1},\lambda_{t}\right)+\frac{1}{2\eta_{t}}d^{2}\left(x,x_{t+1}\right)+\frac{1}{2\eta_{t}}d^{2}\left(x_{t+1},x_{t}\right)\\
&+<s-\eta_{t}exp_{x_{t+1}}^{-1}x_{t},exp_{x_{t+1}}^{-1}x>.\label{eq:13}
\end{split}
\end{equation}

From eq. (\ref{eq:3}), we have 
$0\in\partial L\left(x_{t+1},\lambda_{t}\right)+\eta_{t}exp_{x_{t+1}}^{-1}x_{t}$.
 So

$L\left(x,\lambda_{t}\right)+\frac{1}{2\eta_{t}}d^{2}\left(x,x_{t}\right)\geq L\left(x_{t+1},\lambda_{t}\right)+\frac{1}{2\eta_{t}}d^{2}\left(x,x_{t+1}\right)+\frac{1}{2\eta_{t}}d^{2}\left(x_{t+1},x_{t}\right)$.

Since 
$\frac{1}{2\eta_{t}}d^{2}\left(x_{t+1},x_{t}\right)$ is zero or positive, by taking out it,

$L\left(x_{t+1},\lambda_{t}\right)-L\left(x,\lambda_{t}\right)\leq\frac{1}{2\eta_{t}}d^{2}\left(x,x_{t}\right)-\frac{1}{2\eta_{t}}d^{2}\left(x,x_{t+1}\right)$.

Let 
$\left(x_{*},\lambda_{*}\right)$ be the saddle (min-max) point which satisfies

$L\left(x_{*},\lambda\right)\leq L\left(x_{*},\lambda_{*}\right)\leq L\left(x,\lambda_{*}\right)$. By choosing $x=x_{*}$, we have
\begin{equation}
L\left(x_{t+1},\lambda_{t}\right)-L\left(x_{*},\lambda_{t}\right)\leq\frac{1}{2\eta_{t}}d^{2}\left(x_{*},x_{t}\right)-\frac{1}{2\eta_{t}}d^{2}\left(x_{*},x_{t+1}\right).\label{eq:14}
\end{equation}

Multiplying eq. (\ref{eq:14}) with $\eta_{t}$ and summing over $t=0,1,2,...,T-1$, 
\begin{multline}
\sum_{t=0}^{T-1}\eta_{t}\left(L\left(x_{t+1},\lambda_{t}\right)-L\left(x_{*},\lambda_{t}\right)\right)\\
\leq\frac{1}{2}\sum_{t=0}^{T-1}\left(d^{2}\left(x_{*},x_{t}\right)-d^{2}\left(x_{*},x_{t+1}\right)\right).\label{eq:15}
\end{multline}

For any dual variable 
$\lambda\in\mathbb{R}_{+}^{m}$, we have

$\parallel\lambda_{t+1}-\lambda\parallel^{2}=\parallel\left[\lambda_{t}+\eta_{t}grad_{\lambda_{t}}L\left(x_{t+1},\lambda_{t}\right)\right]_{+}-\lambda\parallel^{2}\leq\parallel\lambda_{t}-\lambda\parallel^{2}+2\eta_{t}<\text{grad}_{\lambda_{t}}L\left(x_{t+1},\lambda_{t}\right),\lambda_{t}-\lambda>+\eta_{t}^{2}\parallel grad_{\lambda_{t}}L\left(x_{t+1},\lambda_{t}\right)\parallel^{2}.$

By summing over 
$t=0,1,2,...,T-1$, and using the telescoping sum series, we have

$\parallel\lambda_{T}-\lambda\parallel^{2}-\parallel\lambda_{0}-\lambda\parallel^{2}\leq\sum_{t=0}^{T-1}\left(2\eta_{t}<\text{grad}_{\lambda_{t}}L\left(x_{t+1},\lambda_{t}\right),\lambda_{t}-\lambda>\right)+\sum_{t=0}^{T-1}\left(\eta_{t}^{2}\parallel grad_{\lambda_{t}}L\left(x_{t+1},\lambda_{t}\right)\parallel^{2}\right)$.

Using the fact that 
$\lambda_{0}=0$ and 
$\parallel\lambda_{T}-\lambda\parallel^{2}\geq0$, we have
\begin{equation}
\begin{split}
&\sum_{t=0}^{T-1}\left(2\eta_{t}<\text{grad}_{\lambda_{t}}L\left(x_{t+1},\lambda_{t}\right),\lambda-\lambda_{t}>\right)\\
&\leq\parallel\lambda\parallel^{2}+\sum_{t=0}^{T-1}\left(\eta_{t}^{2}\parallel grad_{\lambda_{t}}L\left(x_{t+1},\lambda_{t}\right)\parallel^{2}\right).\label{eq:16}
\end{split}
\end{equation}

Because 
$L\left(x_{t+1},\lambda\right)$ is concave with respect to 
$\lambda$, 

$<grad_{\lambda_{t}}L\left(x_{t+1},\lambda_{t}\right),\lambda-\lambda_{t}>\geq L\left(x_{t+1},\lambda\right)-L\left(x_{t+1},\lambda_{t}\right)$.

By replacing 
$<grad_{\lambda_{t}}L\left(x_{t+1},\lambda_{t}\right),\lambda-\lambda_{t}>$ in eq.
 (\ref{eq:16}), we have
\begin{equation}
\begin{split}
&\sum_{t=0}^{T-1}\eta_{t}\left(L\left(x_{t+1},\lambda\right)-L\left(x_{t+1},\lambda_{t}\right)\right)\\
&\leq\frac{1}{2}\parallel\lambda\parallel^{2}+\frac{1}{2}\sum_{t=0}^{T-1}\left(\eta_{t}^{2}\parallel grad_{\lambda_{t}}L\left(x_{t+1},\lambda_{t}\right)\parallel^{2}\right).\label{eq:17}
\end{split}
\end{equation}

Combine eq. (\ref{eq:15}) and eq. (\ref{eq:17}),
\begin{equation}
\begin{split}
&\sum_{t=0}^{T-1}\eta_{t}\left(L\left(x_{t+1},\lambda\right)-L\left(x_{*},\lambda_{t}\right)\right)\\
&\leq\frac{1}{2}\sum_{t=0}^{T-1}\left(d^{2}\left(x_{*},x_{t}\right)-d^{2}\left(x_{*},x_{t+1}\right)\right)+\frac{1}{2}\parallel\lambda\parallel^{2}\\
&+\frac{1}{2}\sum_{t=0}^{T-1}\left(\eta_{t}^{2}\parallel grad_{\lambda_{t}}L\left(x_{t+1},\lambda_{t}\right)\parallel^{2}\right).\label{eq:18}
\end{split}
\end{equation}

To bound the gradient item 
$\parallel grad_{\lambda_{t}}L\left(x_{t+1},\lambda_{t}\right)\parallel^{2}$ in eq.
 (\ref{eq:18}), we employ Lemma 13 from Ref.\cite{khuzani2017stochastic}.
\begin{equation}
\begin{split}
&\sum_{t=0}^{T-1}\eta_{t}\left(L\left(x_{t+1},\lambda\right)-L\left(x_{*},\lambda_{t}\right)\right)\\
&\leq\frac{1}{2}\sum_{t=0}^{T-1}\left(d^{2}\left(x_{*},x_{t}\right)-d^{2}\left(x_{*},x_{t+1}\right)\right)\\
&+\frac{1}{2}\parallel\lambda\parallel^{2}+2mG^{2}\sum_{t=0}^{T-1}\left(\eta_{t}^{2}\right)\\
&\leq\frac{1}{2}d^{2}\left(x_{*},x_{0}\right)+\frac{1}{2}\parallel\lambda\parallel^{2}+2mG^{2}\sum_{t=0}^{T-1}\left(\eta_{t}^{2}\right).\label{eq:20}
\end{split}
\end{equation}

Now let's expand the left side of eq. (\ref{eq:20}), 
\begin{equation*}
\begin{split}
&\sum_{t=0}^{T-1}\eta_{t}(L(x_{t+1},\lambda)-L(x_{*},\lambda_{t}))=\\
&\sum_{t=0}^{T-1}\eta_{t}(f(x_{t+1})+<\lambda,h(x_{t+1})>-\frac{\alpha}{2}\parallel\lambda\parallel^{2}\\
&-(f(x_{*})+<\lambda_{t},h(x_{*})>-\frac{\alpha}{2}\parallel\lambda_{t}\parallel^{2})).
\end{split}
\end{equation*}

Since 
$h\left(x_{*}\right)\preceq0$, $\alpha>0$, 
$\lambda_{t}\succeq0$ and 
$\parallel\lambda_{t}\parallel^{2}\geq0$, by removing positive terms 
$-<\lambda_{t},h\left(x_{*}\right)>$ and 
$\frac{\alpha}{2}\parallel\lambda_{t}\parallel^{2}$, we have the following inequality
\begin{equation}
\begin{split}
&\sum_{t=0}^{T-1}\eta_{t}\left(f\left(x_{t+1}\right)-f\left(x_{*}\right)+<\lambda,h\left(x_{t+1}\right)>-\frac{\alpha}{2}\parallel\lambda\parallel^{2}\right)\\
&\leq\frac{1}{2}d^{2}\left(x_{*},x_{0}\right)+\frac{1}{2}\parallel\lambda\parallel^{2}+2mG^{2}\sum_{t=0}^{T-1}\left(\eta_{t}^{2}\right).\label{eq:21}
\end{split}
\end{equation}

By moving 
$\frac{1}{2}\parallel\lambda\parallel^{2}$ from r.h.s.
 of eq. (\ref{eq:21}) to l.h.s. of eq. (\ref{eq:21}), we have 
\begin{equation}
\begin{split}
&\sum_{t=0}^{T-1}\eta_{t}\left(f\left(x_{t+1}\right)-f\left(x_{*}\right)\right)\\
&+\left(<\lambda,\sum_{t=0}^{T-1}\eta_{t}h\left(x_{t+1}\right)>-\frac{\alpha\left(\sum_{t=0}^{T-1}\eta_{t}\right)+1}{2}\parallel\lambda\parallel^{2}\right)\\
&\leq\frac{1}{2}d^{2}\left(x_{*},x_{0}\right)+2mG^{2}\sum_{t=0}^{T-1}\left(\eta_{t}^{2}\right).\label{eq:22}
\end{split}
\end{equation}

By maximizing 
$\left(<\lambda,\sum_{t=0}^{T-1}\eta_{t}h\left(x_{t+1}\right)>-\frac{\alpha\left(\sum_{t=0}^{T-1}\eta_{t}\right)+1}{2}\parallel\lambda\parallel^{2}\right)$
, we have 
$\lambda_{max}=\left(\left(\alpha\sum_{t=0}^{T-1}\eta_{t}\right)+1\right)^{-1}\left[\sum_{t=0}^{T-1}\eta_{t}h\left(x_{t+1}\right)\right]_{+}$, and 
\begin{equation*}
\begin{split}
&max\left(<\lambda,\sum_{t=0}^{T-1}\eta_{t}h\left(x_{t+1}\right)>-\frac{1}{2}\left(\alpha\left(\sum_{t=0}^{T-1}\eta_{t}\right)+1\right)\parallel\lambda\parallel^{2}\right)\\
&=\frac{1}{2}\left(\left(\alpha\sum_{t=0}^{T-1}\eta_{t}\right)+1\right)^{-1}\parallel\left[\sum_{t=0}^{T-1}\eta_{t}h\left(x_{t+1}\right)\right]_{+}\parallel^{2}.
\end{split}
\end{equation*}

Since 
$\lambda\in\mathbb{\mathbb{R}}_{+}^{m}$ could be any value in 
$R_{+}^{m}$, so
\begin{equation*}
\begin{split}
&\sum_{t=0}^{T-1}\eta_{t}\left(f\left(x_{t+1}\right)-f\left(x_{*}\right)\right)\\
&+\left(<\lambda,\sum_{t=0}^{T-1}\eta_{t}h\left(x_{t+1}\right)>-\frac{\alpha\left(\sum_{t=0}^{T-1}\eta_{t}\right)+1}{2}\parallel\lambda\parallel^{2}\right)\\
&\leq\sum_{t=0}^{T-1}\eta_{t}\left(f\left(x_{t+1}\right)-f\left(x_{*}\right)\right)\\
&+\frac{1}{2}\left(\left(\alpha\sum_{t=0}^{T-1}\eta_{t}\right)+1\right)\parallel\left[\sum_{t=0}^{T-1}\eta_{t}h\left(x_{t+1}\right)\right]_{+}\parallel^{2}\\
&\leq\frac{1}{2}d^{2}\left(x_{*},x_{0}\right)+2mG^{2}\sum_{t=0}^{T-1}\left(\eta_{t}^{2}\right).
\end{split}
\end{equation*}

By removing the maximum term which is positive on the l.h.s. of above equation , we have
\begin{multline*}
\sum_{t=0}^{T-1}\eta_{t}\left(f\left(x_{t+1}\right)-f\left(x_{*}\right)\right)
\leq\frac{1}{2}d^{2}\left(x_{*},x_{0}\right)+2mG^{2}\sum_{t=0}^{T-1}\left(\eta_{t}^{2}\right).
\end{multline*}

Since 
\begin{equation*}
\begin{split}
\sum_{t=0}^{T-1}\eta_{t}\left(f\left(x_{t+1}\right)-f\left(x_{*}\right)\right)
&\geq\sum_{t=0}^{T-1}\eta_{t}\left(\underset{t\in\left[T\right]}{min}f\left(x_{t+1}\right)-f\left(x_{*}\right)\right)\\
&=\left(\underset{t\in\left[T\right]}{min}f\left(x_{t+1}\right)-f\left(x_{*}\right)\right)\sum_{t=0}^{T-1}\eta_{t},
\end{split}
\end{equation*}
we have
\begin{equation}
\begin{split}
&\left(\underset{t\in\left[T\right]}{min}f\left(x_{t+1}\right)-f\left(x_{*}\right)\right)\\
&\leq\frac{1}{\sum_{t=0}^{T-1}\eta_{t}}\left(\frac{1}{2}d^{2}\left(x_{*},x_{0}\right)+2mG^{2}\sum_{t=0}^{T-1}\left(\eta_{t}^{2}\right)\right).\label{eq:24}
\end{split}
\end{equation}
End of proof.

\subsection{Proof of Corollary 1}
The proof simply follows the proof of Corollary 8 in ref.\cite{khuzani2017stochastic}.

First we have the following bounds: 
\begin{align*}
\sum_{t=0}^{T-1}\left(\eta_{t}\right)=&\sum_{t=0}^{T-1}\frac{1}{\sqrt{t+1}}\geq\int_{0}^{T-1}\frac{1}{\sqrt{t+1}}dt\\
=&2\sqrt{t+1}\mid_{t=0}^{T-1}=2\left(\sqrt{T}-1\right),
\end{align*}
\begin{align*}
\sum_{t=0}^{T-1}\left(\eta_{t}^{2}\right)=\sum_{t=0}^{T-1}\frac{1}{t+1}\leq1+\int_{0}^{T-1}\frac{1}{t+1}dt=1+log\left(T\right). 
\end{align*}
By using the fact that $d^{2}\left(x_{*},x_{0}\right)\leq R^{2}$, we have 
\begin{multline*}
\left(\underset{t\in\left[T\right]}{min}f\left(x_{t+1}\right)-f\left(x_{*}\right)\right)\leq\\
\frac{1}{2\left(\sqrt{T}-1\right)}\left(\frac{1}{2}R^{2}+2mG^{2}\left(1+log\left(T\right)\right)\right)=\mathcal{O}\left(\frac{log\left(T\right)}{\sqrt{T}-1}\right).
\end{multline*}
End of proof.

\subsection{Riemannian Gradient with Retraction}
We can write Euclidean gradient as 
$grad_{\mathbf{W}}J_{\mathbf{W}}$

$=\frac{\partial}{\partial \mathbf{W}}\left(\frac{1}{2}d^{2}\left(\mathbf{W}_{t},\mathbf{W}\right)+<\lambda_{t},h\left(\mathbf{W}\right)>\right)$

$=\frac{1}{2}\left(\mathbf{W}_{0}^{-1}-\mathbf{W}^{-1}\right)+<\lambda_{t},\frac{\partial h\left(\mathbf{W}\right)}{\partial \mathbf{W}}>$

$\frac{\partial h\left(\mathbf{W}\right)}{\partial \mathbf{W}}=\frac{\partial}{\partial \mathbf{W}}\left(\left[h_{+}\left(\mathbf{W}\right),h_{-}\left(\mathbf{W}\right)\right]^{\top}\right)$
$=\left[diag\left(Diag\left(\mathbf{X}_{+}^{\top}\right)Diag\left(\mathbf{X}_{+}\right)\right),-diag\left(Diag\left(\mathbf{X}_{-}^{\top}\right)Diag\left(\mathbf{X}_{-}\right)\right)\right]^{\top}$, 
where 
$Diag\left(\mathbf{X}_{+}^{\top}\right) / Diag\left(\mathbf{X}_{+}\right)$ means constructing a block diagonal matrix whose block diagonal elements are columns / rows of $\mathbf{X}_{+}^{\top} / \mathbf{X}_{+}$. In another word, each element of 
$\frac{\partial h\left(\mathbf{W}\right)}{\partial \mathbf{W}}$ is a covariance matrix 
$\mathbf{x}_{i}^{\top}\times\mathbf{x}_{i}$ of a sample pair vector $\mathbf{x}_{i}$, $i=1,2,...,N$ (assume the expectation of $\mathbf{x}_{i}$, $i=1,2,...,N$, is zero).
 
$<\lambda_{t},\frac{\partial h\left(\mathbf{W}\right)}{\partial \mathbf{W}}>$ is tensor dot product between $\lambda_{t}$ and a vector of size $N$ whose element is covariance matrix of the $i$'th sample pair.

\textbf{1) Riemannian Gradient}

First we show that in our case, $Grad_{\mathbf{W}}J_{\mathbf{W}}=grad_{\mathbf{W}}J_{\mathbf{W}}$.

Let's define 
$M_{W}:=\left\{ W=U\Sigma V^{\top}\right\} $, s.t. $U\in st\left(m,d\right)$, $V\in st\left(n,d\right)$, and $\Sigma=diag\left(\sigma_{1},\sigma_{2},...,\sigma_{d}\right)$, with $\sigma_{i}>0$, $\forall i$. $st\left(m,d\right)$ is Stiefel manifold of $m\times d$ real, orthonormal matrices. $M_{W}$ is a Riemannian manifold with tangent space\cite{absil2009optimization,Vandereycken2013}:

$T_{W}M:=\left\{ UMV^{\top}+U_{p}V^{\top}+UV_{p}^{\top}\right\} $, where 
$M$ is an arbitrary $d\times d$ matrix, $U_{p}^{\top}U=0$, and $V_{p}^{\top}V=0$.
 
For a given objective function 
$J$ which depends on input matrix W of size 
$m\times n$, we use 
$grad_{\mathbf{W}}J\in\mathbb{R}^{m\times n}$ represent Euclidean gradient of 
$J$ w.r.t $W$; and denote 
$Grad_{\mathbf{W}}J$ as its Riemannian gradient by projecting the Euclidean gradient onto the
 tangent space of $M_{W}$:

$Grad_{\mathbf{W}}J=P_{U}^{H}grad_{\mathbf{W}}JP_{V}^{H}+P_{U}^{\upsilon}grad_{\mathbf{W}}JP_{V}^{H}+P_{U}^{H}grad_{\mathbf{W}}JP_{V}^{\upsilon}$, where 
$P_{U}^{H}:=UU^{\top}$, $P_{U}^{\upsilon}:=I-UU^{\top}$, $P_{V}^{H}:=VV^{\top}$, and $P_{V}^{\upsilon}:=I-VV^{\top}$.

For our problem, since $W$ is a real symmetric positive definite matrix, we have $M_{W}:=\left\{ W=Q\Lambda Q^{\top}\right\} $, and $T_{W}M:=\left\{ QMQ^{\top}\right\} $, where $Q$ is an orthogonal matrix, $\Lambda$ is a diagonal matrix whose entries are the eigenvalues of $W$ and greater than or equal to zero.

Since 
$P_{U}^{H}:=UU^{\top}=QQ^{\top}=I$, 
$P_{U}^{\upsilon}:=I-UU^{\top}=0$, 
$P_{V}^{H}:=VV^{\top}=I$, 
$P_{V}^{\upsilon}:=I-VV^{\top}=0$
, the projection of the Euclidean gradient 
$grad_{\mathbf{W}}J_{\mathbf{W}}$ onto the tangent space of $M_{W}$ is

$Grad_{\mathbf{W}}J_{\mathbf{W}}=P_{U}^{H}grad_{\mathbf{W}}J_{\mathbf{W}}P_{V}^{H}+P_{U}^{\upsilon}grad_{\mathbf{W}}J_{\mathbf{W}}P_{V}^{H}+P_{U}^{H}grad_{\mathbf{W}}J_{\mathbf{W}}P_{V}^{\upsilon}=grad_{\mathbf{W}}J_{\mathbf{W}}$.

\textbf{2) Retraction}

With 
$\mathbf{W}_{t}$
 and 
$Grad_{\mathbf{W}}J_{\mathbf{W}}$
 shown above, we would like to calculate 
$\mathbf{W}_{t+1}$ using retraction.
 From Ref.\cite{cheng2013riemannian}, for any tangent vector 
$\eta\in T_{W}M$, its retraction 
$R_{W}\left(\eta\right):=\underset{X\in M}{armin}\parallel W+\eta-X\parallel_{F}$.
 For our case $R_{\mathbf{W}_{t}}\left(-\eta_{t}Grad_{\mathbf{W}_{t}}J_{\mathbf{W}_{t}}\right)=\sum_{i=1}^n\sigma_{i}q_{i}q_{i}^{\top}$
, where $\sigma_{i}$ and $q_{i}$ are the $i$-th eigenvalues and eigenvector of matrix 
$\mathbf{W}_{t}-\eta_{t}Grad_{\mathbf{W}_{t}}J_{\mathbf{W}_{t}}$.

The calculation and retraction shown above are repeated until 
$J_{\mathbf{W}}$ converges.

\bibliographystyle{IEEEtran}
\bibliography{riemannian_IJCNN}

\begin{thebibliography}{10}
\providecommand{\url}[1]{#1}
\csname url@samestyle\endcsname
\providecommand{\newblock}{\relax}
\providecommand{\bibinfo}[2]{#2}
\providecommand{\BIBentrySTDinterwordspacing}{\spaceskip=0pt\relax}
\providecommand{\BIBentryALTinterwordstretchfactor}{4}
\providecommand{\BIBentryALTinterwordspacing}{\spaceskip=\fontdimen2\font plus
\BIBentryALTinterwordstretchfactor\fontdimen3\font minus
  \fontdimen4\font\relax}
\providecommand{\BIBforeignlanguage}[2]{{%
\expandafter\ifx\csname l@#1\endcsname\relax
\typeout{** WARNING: IEEEtran.bst: No hyphenation pattern has been}%
\typeout{** loaded for the language `#1'. Using the pattern for}%
\typeout{** the default language instead.}%
\else
\language=\csname l@#1\endcsname
\fi
#2}}
\providecommand{\BIBdecl}{\relax}
\BIBdecl

\bibitem{nishimori2006riemannian}
Y.~Nishimori, S.~Akaho, and M.~D. Plumbley, ``Riemannian optimization method on
  the flag manifold for independent subspace analysis,'' in \emph{International
  Conference on Independent Component Analysis and Signal Separation}.\hskip
  1em plus 0.5em minus 0.4em\relax Springer, 2006, pp. 295--302.

\bibitem{absil2007trust}
P.-A. Absil, C.~G. Baker, and K.~A. Gallivan, ``Trust-region methods on
  riemannian manifolds,'' \emph{Foundations of Computational Mathematics},
  vol.~7, no.~3, pp. 303--330, 2007.

\bibitem{Vandereycken2013}
B.~Vandereycken, ``{Low-rank matrix completion by Riemannian optimization},''
  \emph{SIAM Journal on Optimization}, vol. 23, No. 2, pp. 1214--1236, 2013.

\bibitem{cheng2013riemannian}
L.~Cheng, ``Riemannian similarity learning,'' in \emph{International Conference
  on Machine Learning}, 2013, pp. 540--548.

\bibitem{cherian2017riemannian}
A.~Cherian and S.~Sra, ``Riemannian dictionary learning and sparse coding for
  positive definite matrices,'' \emph{IEEE transactions on neural networks and
  learning systems}, vol.~28, no.~12, pp. 2859--2871, 2017.

\bibitem{absil2010optimization}
P.-A. Absil, R.~Mahony, and R.~Sepulchre, ``Optimization on manifolds: Methods
  and applications,'' in \emph{Recent Advances in Optimization and its
  Applications in Engineering}.\hskip 1em plus 0.5em minus 0.4em\relax
  Springer, 2010, pp. 125--144.

\bibitem{gabay1982minimizing}
D.~Gabay, ``Minimizing a differentiable function over a differential
  manifold,'' \emph{Journal of Optimization Theory and Applications}, vol.~37,
  no.~2, pp. 177--219, 1982.

\bibitem{smith1994optimization}
S.~T. Smith, ``Optimization techniques on riemannian manifolds,'' \emph{Fields
  institute communications}, vol.~3, no.~3, pp. 113--135, 1994.

\bibitem{BentoFO12}
G.~de~Carvalho~Bento, O.~P. Ferreira, and P.~R. Oliveira, ``Unconstrained
  steepest descent method for multicriteria optimization on riemannian
  manifolds,'' \emph{J. Optimization Theory and Applications}, vol. 154, no.~1,
  pp. 88--107, 2012.

\bibitem{sato2015new}
H.~Sato and T.~Iwai, ``A new, globally convergent riemannian conjugate gradient
  method,'' \emph{Optimization}, vol.~64, no.~4, pp. 1011--1031, 2015.

\bibitem{liu2017accelerated}
Y.~Liu, F.~Shang, J.~Cheng, H.~Cheng, and L.~Jiao, ``Accelerated first-order
  methods for geodesically convex optimization on riemannian manifolds,'' in
  \emph{Advances in Neural Information Processing Systems}, 2017, pp.
  4868--4877.

\bibitem{qi2010riemannian}
C.~Qi, K.~A. Gallivan, and P.-A. Absil, ``Riemannian bfgs algorithm with
  applications,'' in \emph{Recent advances in optimization and its applications
  in engineering}.\hskip 1em plus 0.5em minus 0.4em\relax Springer, 2010, pp.
  183--192.

\bibitem{abs-1810-00760}
\BIBentryALTinterwordspacing
G.~B{\'{e}}cigneul and O.~Ganea, ``Riemannian adaptive optimization methods,''
  \emph{CoRR}, vol. abs/1810.00760, 2018. [Online]. Available:
  \url{http://arxiv.org/abs/1810.00760}
\BIBentrySTDinterwordspacing

\bibitem{qi2011numerical}
C.~Qi, ``Numerical optimization methods on riemannian manifolds,'' 2011.

\bibitem{agarwal2018adaptive}
N.~Agarwal, N.~Boumal, B.~Bullins, and C.~Cartis, ``Adaptive regularization
  with cubics on manifolds with a first-order analysis,'' \emph{arXiv preprint
  arXiv:1806.00065}, 2018.

\bibitem{schmidt2017minimizing}
M.~Schmidt, N.~Le~Roux, and F.~Bach, ``Minimizing finite sums with the
  stochastic average gradient,'' \emph{Mathematical Programming}, vol. 162, no.
  1-2, pp. 83--112, 2017.

\bibitem{johnson2013accelerating}
R.~Johnson and T.~Zhang, ``Accelerating stochastic gradient descent using
  predictive variance reduction,'' in \emph{Advances in neural information
  processing systems}, 2013, pp. 315--323.

\bibitem{defazio2014saga}
A.~Defazio, F.~Bach, and S.~Lacoste-Julien, ``Saga: A fast incremental gradient
  method with support for non-strongly convex composite objectives,'' in
  \emph{Advances in neural information processing systems}, 2014, pp.
  1646--1654.

\bibitem{zhang2016riemannian}
H.~Zhang, S.~J. Reddi, and S.~Sra, ``Riemannian svrg: Fast stochastic
  optimization on riemannian manifolds,'' in \emph{Advances in Neural
  Information Processing Systems}, 2016, pp. 4592--4600.

\bibitem{sato2017riemannian}
H.~Sato, H.~Kasai, and B.~Mishra, ``Riemannian stochastic variance reduced
  gradient,'' \emph{arXiv preprint arXiv:1702.05594}, 2017.

\bibitem{kasai2017riemannian}
H.~Kasai, H.~Sato, and B.~Mishra, ``Riemannian stochastic quasi-newton
  algorithm with variance reduction and its convergence analysis,'' \emph{arXiv
  preprint arXiv:1703.04890}, 2017.

\bibitem{hauswirth2016projected}
A.~Hauswirth, S.~Bolognani, G.~Hug, and F.~D{\"o}rfler, ``Projected gradient
  descent on riemannian manifolds with applications to online power system
  optimization,'' in \emph{2016 54th Annual Allerton Conference on
  Communication, Control, and Computing (Allerton)}.\hskip 1em plus 0.5em minus
  0.4em\relax IEEE, 2016, pp. 225--232.

\bibitem{zhang2017primal}
J.~Zhang, S.~Ma, and S.~Zhang, ``Primal-dual optimization algorithms over
  riemannian manifolds: an iteration complexity analysis,'' \emph{arXiv
  preprint arXiv:1710.02236}, 2017.

\bibitem{khuzani2017stochastic}
M.~B. Khuzani and N.~Li, ``Stochastic primal-dual method on riemannian
  manifolds of bounded sectional curvature,'' in \emph{Machine Learning and
  Applications (ICMLA), 2017 16th IEEE International Conference on}.\hskip 1em
  plus 0.5em minus 0.4em\relax IEEE, 2017, pp. 133--140.

\bibitem{rockafellar1976augmented}
R.~T. Rockafellar, ``Augmented lagrangians and applications of the proximal
  point algorithm in convex programming,'' \emph{Mathematics of operations
  research}, vol.~1, no.~2, pp. 97--116, 1976.

\bibitem{parikh2014proximal}
N.~Parikh, S.~Boyd \emph{et~al.}, ``Proximal algorithms,'' \emph{Foundations
  and Trends{\textregistered} in Optimization}, vol.~1, no.~3, pp. 127--239,
  2014.

\bibitem{ferreira2002proximal}
O.~Ferreira and P.~Oliveira, ``Proximal point algorithm on riemannian
  manifolds,'' \emph{Optimization}, vol.~51, no.~2, pp. 257--270, 2002.

\bibitem{bavcak2013proximal}
M.~Ba{\v{c}}{\'a}k, ``The proximal point algorithm in metric spaces,''
  \emph{Israel Journal of Mathematics}, vol. 194, no.~2, pp. 689--701, 2013.

\bibitem{chen2018proximal}
S.~Chen, S.~Ma, A.~M.-C. So, and T.~Zhang, ``Proximal gradient method for
  manifold optimization,'' \emph{arXiv preprint arXiv:1811.00980}, 2018.

\bibitem{eisenhart2016riemannian}
L.~P. Eisenhart, \emph{Riemannian geometry}.\hskip 1em plus 0.5em minus
  0.4em\relax Princeton university press, 2016.

\bibitem{kulis2013metric}
B.~Kulis \emph{et~al.}, ``Metric learning: A survey,'' \emph{Foundations and
  Trends{\textregistered} in Machine Learning}, vol.~5, no.~4, pp. 287--364,
  2013.

\bibitem{bellet2013survey}
A.~Bellet, A.~Habrard, and M.~Sebban, ``A survey on metric learning for feature
  vectors and structured data,'' \emph{arXiv preprint arXiv:1306.6709}, 2013.

\bibitem{xing2003distance}
E.~P. Xing, M.~I. Jordan, S.~J. Russell, and A.~Y. Ng, ``Distance metric
  learning with application to clustering with side-information,'' in
  \emph{Advances in neural information processing systems}, 2003, pp. 521--528.

\bibitem{weinberger2006distance}
K.~Q. Weinberger, J.~Blitzer, and L.~K. Saul, ``Distance metric learning for
  large margin nearest neighbor classification,'' in \emph{Advances in neural
  information processing systems}, 2006, pp. 1473--1480.

\bibitem{davis2007information}
J.~V. Davis, B.~Kulis, P.~Jain, S.~Sra, and I.~S. Dhillon,
  ``Information-theoretic metric learning,'' in \emph{Proceedings of the 24th
  international conference on Machine learning}.\hskip 1em plus 0.5em minus
  0.4em\relax ACM, 2007, pp. 209--216.

\bibitem{wang2009information}
S.~Wang and R.~Jin, ``An information geometry approach for distance metric
  learning,'' in \emph{Artificial Intelligence and Statistics}, 2009, pp.
  591--598.

\bibitem{zadeh2016geometric}
P.~Zadeh, R.~Hosseini, and S.~Sra, ``Geometric mean metric learning,'' in
  \emph{International Conference on Machine Learning}, 2016, pp. 2464--2471.

\bibitem{mahadevan2018unified}
S.~Mahadevan, B.~Mishra, and S.~Ghosh, ``A unified framework for domain
  adaptation using metric learning on manifolds,'' \emph{arXiv preprint
  arXiv:1804.10834}, 2018.

\bibitem{levitin1966constrained}
E.~S. Levitin and B.~T. Polyak, ``Constrained minimization methods,''
  \emph{Zhurnal Vychislitel'noi Matematiki i Matematicheskoi Fiziki}, vol.~6,
  no.~5, pp. 787--823, 1966.

\bibitem{shalev2004online}
S.~Shalev-Shwartz, Y.~Singer, and A.~Y. Ng, ``Online and batch learning of
  pseudo-metrics,'' in \emph{Proceedings of the twenty-first international
  conference on Machine learning}.\hskip 1em plus 0.5em minus 0.4em\relax ACM,
  2004, p.~94.

\bibitem{jain2009online}
P.~Jain, B.~Kulis, I.~S. Dhillon, and K.~Grauman, ``Online metric learning and
  fast similarity search,'' in \emph{Advances in neural information processing
  systems}, 2009, pp. 761--768.

\bibitem{defusco2015quantitative}
R.~A. DeFusco, D.~W. McLeavey, J.~E. Pinto, M.~J. Anson, and D.~E. Runkle,
  \emph{Quantitative investment analysis}.\hskip 1em plus 0.5em minus
  0.4em\relax John Wiley \& Sons, 2015.

\bibitem{fama2004capital}
E.~F. Fama and K.~R. French, ``The capital asset pricing model: Theory and
  evidence,'' \emph{Journal of economic perspectives}, vol.~18, no.~3, pp.
  25--46, 2004.

\bibitem{bento2017iteration}
G.~C. Bento, O.~P. Ferreira, and J.~G. Melo, ``Iteration-complexity of
  gradient, subgradient and proximal point methods on riemannian manifolds,''
  \emph{Journal of Optimization Theory and Applications}, vol. 173, no.~2, pp.
  548--562, 2017.

\bibitem{absil2009optimization}
P.~A. Absil, R.~Mahony, and R.~Sepulchre, \emph{Optimization algorithms on
  matrix manifolds}.\hskip 1em plus 0.5em minus 0.4em\relax Princeton
  University Press, 2009.

\end{thebibliography}

\end{document}